\documentclass{article}
\PassOptionsToPackage{numbers,compress,sort}{natbib}
\usepackage[final,sglblindworkshop]{neurips_2026}
\workshoptitle{Second Workshop on ML$\times$OR}

\usepackage[utf8]{inputenc}
\usepackage[T1]{fontenc}
\usepackage{amsmath}
\usepackage{amsfonts}
\usepackage{booktabs}
\usepackage{float}
\usepackage{graphicx}
\usepackage{hyperref}
\usepackage{microtype}
\usepackage{url}
\usepackage{xcolor}
\hypersetup{hidelinks}

\title{RELAGRAM:
  \texorpdfstring{\underline{Re}cursive \underline{La}tent
  \underline{Gra}ph \underline{M}odel}{Recursive Latent Graph Model}
  for the Capacitated Vehicle Routing Problem}

\author{%
\begin{minipage}[t]{0.47\textwidth}
\normalfont
\centering
\textbf{Vladimir Kirill Bickov}\\
Delft University of Technology, \\ The Netherlands\\
\texttt{me@nevolodia.com}
\end{minipage}
\hfill
\begin{minipage}[t]{0.47\textwidth}
\normalfont
\centering
\textbf{Neil Yorke-Smith}\\
Delft University of Technology, \\ The Netherlands\\
\texttt{n.yorke-smith@tudelft.nl}
\end{minipage}%
}

\begin{document}

\maketitle

\begin{abstract}
Latent recursion repeatedly processes internal
representations through the same neural network.
Recent work has shown this approach to be
effective on structured reasoning tasks such
as Sudoku and maze solving.  However, whether
latent recursion can support neural routing
remains unclear.
We introduce RELAGRAM (\underline{Re}cursive
\underline{La}tent \underline{Gra}ph
\underline{M}odel), a novel approach that
brings latent recursion to graph combinatorial
optimization.  RELAGRAM encodes a capacitated
vehicle routing problem (CVRP) instance as
edge tokens; a compact transformer recurrently
updates the latent representations of these tokens.  An
output head maps the answer state to an edge
heatmap, from which an autoregressive decoder
constructs a feasible solution.  On CVRP-100 instances,
the 0.53M-parameter RELAGRAM achieves gaps
to a PyVRP reference of 13.4\%, 9.8\%, and 8.3\%
under greedy decoding, eightfold symmetry
augmentation, and 256-candidate search,
respectively.  Trained only on CVRP-100, RELAGRAM
generalizes zero-shot to CVRP-500 and
CVRP-1000, achieving gaps of 15.6\% and 21.7\%
under 256-candidate search.  Together, these
results demonstrate that latent recursion is
a promising approach to neural routing.
\end{abstract}

\section{Introduction}
\label{sec:intro}

Solving a combinatorial optimization problem in a single,
end-to-end pass of a neural network is inherently difficult.
A learned model must navigate a constrained and exponentially
large solution space with limited computation. Increasing
network depth provides more computation, but also raises
training and inference costs~\cite{bengio2021machine}.

Neural-based routing methods commonly construct routes autoregressively
or iteratively refine complete solutions~\cite{kool2019attention,
chen2019neurewriter,wu2022improvement}. Both approaches simplify
the task by breaking the solution process into a series of simpler
decisions. However, this simplicity comes at the cost of an
information bottleneck. If only the updated solution is carried
between iterations, information that exists only in the latent
state, such as intermediate reasoning, is lost. This creates a
form of representational `amnesia' that makes it difficult to
accumulate a richer understanding of the problem.

\begin{figure}[t]
  \centering
  \includegraphics[width=\linewidth]{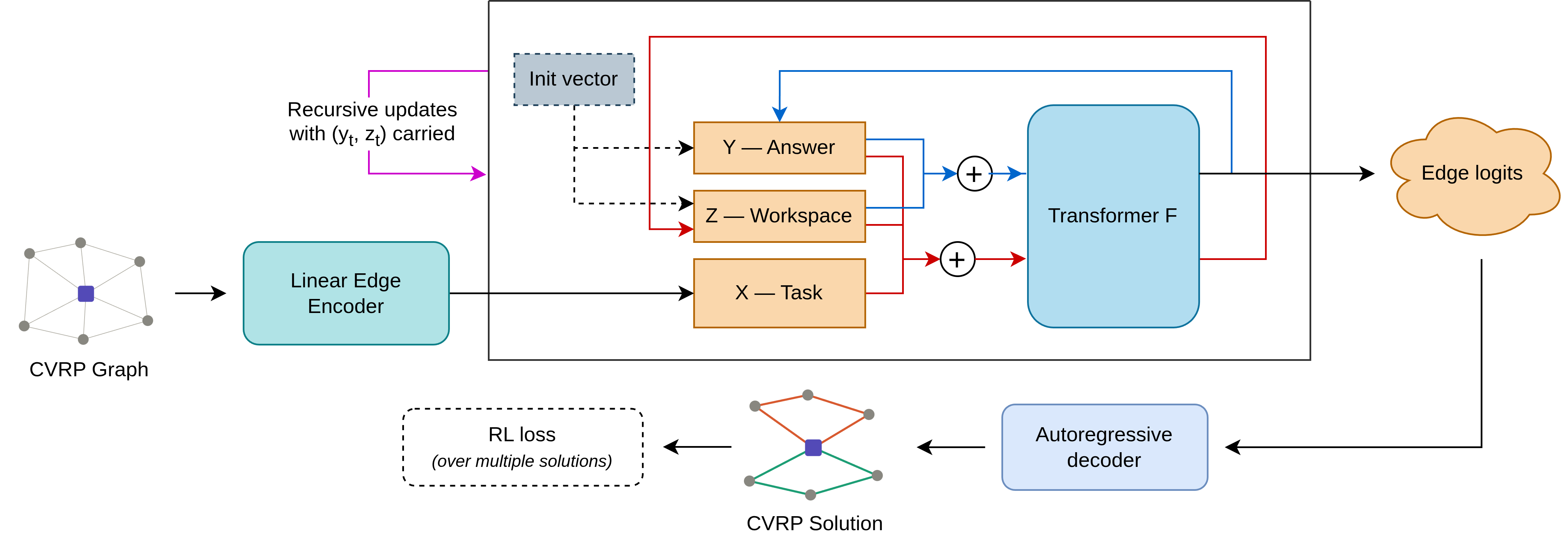}
  
  \caption{Overview of RELAGRAM.  A CVRP instance is encoded as edge-token input $x$;
  the answer state $y$ and workspace $z$ start from fixed vectors.
  The shared transformer first updates $z$ from $x+y+z$ (red), then 
  updates $y$ from $y+z$ (blue).
  Both states $(y,z)$ are carried forward (magenta) in a recursive loop.
  A linear head maps $y_t$ to edge logits, which are decoded into
  a solution.}

  \label{fig:relagram-pipeline}
\end{figure}

Latent recursion avoids this bottleneck by carrying the internal state, rather
than the candidate solution, between iterations. A solution can be decoded
at any stage but is not fed back into the model: each subsequent stage
continues directly from the preceding latent state. Every depth therefore
produces both an intermediate computation and a valid solution point.
Recent latent-recursive models perform strongly on Sudoku, maze solving, and
ARC-AGI~\cite{jolicoeurmartineau2025trm}.

However, whether latent recursion transfers from supervised reasoning to
reinforcement learning for graph combinatorial optimization remains unclear.

We introduce RELAGRAM, a novel application of the latent recursion idea
to graph combinatorial optimization, and demonstrate it on CVRP.
After brief training, a 0.53M model reaches a 13.4\% greedy gap on
CVRP-100, reducing it to 8.3\% under a
256-candidate search.

On CVRP-1000, RELAGRAM achieves gaps of 29.9\% and 21.7\% under greedy and
256-candidate decoding, respectively. Notably, it continues to improve
beyond its trained depth: on CVRP-1000 it achieves its best greedy gap of
28.45\% at depth ten --- six stages deeper than seen during training.

\section{Background}

\paragraph{Latent recursion.}
Latent recursion applies the same network repeatedly to an internal state.
The Tiny Recursive Model (TRM) keeps two states: a solution state $y$ and a latent
reasoning state $z$. One cycle refines $z$ over repeated passes conditioned on
the input and on $y$, then updates $y$ from the refined $z$; a supervision step
chains several such cycles, all but the last without gradients. Deep
supervision gives every step its own loss, and the states are detached
between steps, so the model learns successive corrections without
backpropagating across them. TRM achieves strong results on Sudoku,
maze solving, and ARC-AGI benchmarks~\cite{jolicoeurmartineau2025trm}.

\paragraph{Edge heatmaps.}
An edge heatmap records the model's confidence that each transition belongs to
a high-quality solution. DIFUSCO produces such heatmaps for the travelling
salesman problem by iteratively denoising edge variables~\cite{sun2023difusco}.
However, DIFUSCO's learned edge representation is
separate from the procedure that converts it into a discrete solution.

\paragraph{Feasible decoding.}
A heatmap is not itself a route. For CVRP, each decision must account for the
visited customers and remaining vehicle capacity. The Attention Model, a
neural routing model, constructs a route autoregressively, selecting one node
at a time while masking infeasible actions~\cite{kool2019attention}. DeepACO
instead learns a heuristic measure for an ant colony optimization
system: a population of simulated ants builds routes in parallel, each choosing
only feasible transitions~\cite{ye2023deepaco}.

\paragraph{Learning from sampled routes.}
REINFORCE allows a routing policy to learn without optimal labels: complete
routes are sampled, scored by their cost, and used to improve the
policy~\cite{williams1992reinforce,kool2019attention}. The signal is
sparse, however, because one route-level reward must guide every edge decision
made during construction. Self-imitation adds denser feedback: the lowest-cost
sampled route serves as a temporary teacher, and a cross-entropy loss trains
the heatmap to favour its transitions~\cite{pirnay2024self}.

\section{RELAGRAM}

RELAGRAM brings latent recursion to routing. Between depths it passes its
internal state rather than the route itself, so intermediate reasoning is never
discarded (Figure~\ref{fig:relagram-pipeline}).

\paragraph{Directed edge representation.}
RELAGRAM encodes a CVRP instance as sparse directed edge tokens. Each customer
$i$ connects to the depot and its $k-1$ nearest customers, while the depot
connects to every customer; opposite directions are encoded independently.
For each edge $e=(i,j)$, the encoder collects nine features: the coordinates,
demands, and depot indicators of $i$ and $j$, and their Euclidean distance.
A bias-free linear layer projects them to $x_e\in\mathbb{R}^d$ and scales by
$\sqrt{d}$.

\paragraph{Recursion over edge tokens.}
RELAGRAM maintains answer and workspace states $y_t,z_t$, initialized as
fixed random vectors $y_0,z_0\in\mathbb{R}^d$. The network $F_\theta$ is a
two-block transformer, each block applying
self-attention and SwiGLU~\cite{shazeer2020glu} sublayers with post-norm
residuals and RMSNorm~\cite{zhang2019rmsnorm} without learned scale, as in TRM.
At each depth, it updates $z$ then $y$:
\begin{equation}
z_t=F_\theta(z_{t-1}+y_{t-1}+x), \qquad
y_t=F_\theta(y_{t-1}+z_t).
\label{eq:recurrence}
\end{equation}
A bias-free linear head maps $y_{t,e}$ to $H_{t,ij}$, the score of moving
from $i$ to $j$.
Edges absent from the sparse encoding receive a logit of
$-10$, providing a fallback when every encoded move is masked.

\paragraph{Route decoding.}
Starting at the depot, the decoder constructs a route vertex by vertex. The
row of $H_t$ scores each move after masking visited and
capacity-infeasible customers. Stochastic decoding samples the remaining
logits; greedy takes the largest. The heatmap remains fixed.

\paragraph{Learning and the self-imitation auxiliary.}
At each training depth $t$, RELAGRAM draws $S=16$ routes from the masked
autoregressive policy. Route $\tau_{t,s}$ receives reward
$R_{t,s}=-C(\tau_{t,s})$; their mean $\bar R_t=S^{-1}\sum_sR_{t,s}$ is the
same-instance REINFORCE baseline~\cite{williams1992reinforce}. The loss is
\begin{equation}
\mathcal L_t=
\underbrace{-\frac{1}{S}\sum_{s=1}^{S}(R_{t,s}-\bar R_t)
\log\pi_\Theta(\tau_{t,s}\mid H_t)}_{\text{REINFORCE}}
+\underbrace{\lambda_{\rm SI}\,\mathcal L_{\rm edge}
(H_t,\tau_t^{\rm elite})}_{\text{Self-imitation}}.
\label{eq:loss}
\end{equation}
REINFORCE favours routes shorter than the same-instance mean. Self-imitation
pulls the unmasked heatmap toward the shortest sampled route
$\tau_t^{\rm elite}$ using cross-entropy loss.
We set $\lambda_{\rm SI}=0.3$; $\Theta$ represents trainable parameters.
Every recursion depth receives
one optimizer update, after which its latent states are detached,
so gradients never propagate down the recursive chain.

\section{Experiments}

\paragraph{Experimental design.}
We train on CVRP-100 in the RL4CO \texttt{CVRPEnv}~\cite{berto2025rl4co}, whose
generator samples fresh unit-square instances for every batch.
We use a neighbour mask of $k=20$.
The model has 0.53M parameters and is trained at four depths.
Runs use a batch size of 16 with 16 sampled routes per instance, which
across the four recursion depths yields 1{,}024 routes per batch.
We evaluate RELAGRAM under two training budgets: 40{,}000 optimizer
updates with a training time of 2.98 hours to compare RELAGRAM with
DeepACO; and 160{,}000 optimizer updates with a training time of
11.94 hours to assess performance and its scaling under a larger
training budget.
Training uses a 10\,GB A100 GPU slice.
DeepACO is chosen as our baseline because it is a competitive neural
combinatorial optimizer with a similar design: a learned edge heatmap guides
a separate solution search. DeepACO iterates through pheromone updates,
whereas RELAGRAM updates latent states.
We train it as a 0.52M-parameter model with the same
$k=20$ neighbour mask and optimizer updates matched to RELAGRAM's 40{,}000. The
models use similar numbers of training trajectories (9.60M for DeepACO and
10.24M for RELAGRAM), but the comparison is not compute-matched: DeepACO sees
twice as many base graphs and takes 1.87 times longer to train.
Every model is evaluated on the same 256 fixed instances per size.
References are PyVRP 0.13.4~\cite{wouda2024pyvrp} under per-instance time
limits of 5\,s at $n=100$, 20\,s at $n=500$, and 40\,s at $n=1000$.
We evaluate three decoding configurations: greedy (one route),
eightfold symmetry augmentation, and the 256-candidate combination
of symmetry, per-step latent noise~\cite{sghaier2026ptrm}, and
forced different starts.

\paragraph{Solution quality across graph sizes.}
We evaluate RELAGRAM trained for 160{,}000 updates on CVRP-100.
The model reaches an 8.30\% gap on CVRP-100 under 256-candidate decoding
(Table~\ref{tab:main}). Without retraining, the model achieves gaps of
15.64\% and 21.65\% on CVRP-500 and CVRP-1000 under 256-candidate decoding.
On CVRP-1000,
expanding one greedy decode to 256 candidates improves the gap from 29.91\%
to 21.65\%. The decoder therefore offers test-time scaling that strengthens
with instance size, worth 5.10 points on CVRP-100 and 8.26 on CVRP-1000.

\paragraph{Comparison with DeepACO\@.}
We compare RELAGRAM with DeepACO after training both for 40{,}000 optimizer
updates. DeepACO is stronger in eight of nine size--decoder regimes: it
achieves 3.94-point and 3.49-point lower gaps on CVRP-100 under
greedy and 256-candidate decoding, and a 9.33-point lower greedy gap on
CVRP-1000. RELAGRAM does not yet achieve solution
quality competitive with DeepACO. Under 256 candidates, the two converge as
instances grow: ants sample every step, so DeepACO's candidate scaling weakens
with size while RELAGRAM's does not.

\begin{table}[t]
\centering\small
\caption{Gap to the PyVRP reference (\%; lower is better). For RELAGRAM,
Combined256 uses symmetry, noise, and starts; DeepACO uses 16 ants for
16 iterations.}
\label{tab:main}
{\fontsize{7.5}{8.5}\selectfont
\begin{tabular}{llrrr}
\toprule
& & RELAGRAM & RELAGRAM & DeepACO \\
& decoder & 160k & 40k & 40k \\
\midrule
CVRP-100  & greedy      & 13.40 & 15.55 & 11.61 \\
          & aug8        &  9.77 & 11.30 &  9.16 \\
          & combined256 &  8.30 &  9.57 &  6.08 \\
\midrule
CVRP-500  & greedy      & 22.55 & 24.78 & 19.88 \\
          & aug8        & 17.61 & 19.30 & 17.68 \\
          & combined256 & 15.64 & 17.11 & 17.38 \\
\midrule
CVRP-1000 & greedy      & 29.91 & 33.24 & 23.91 \\
          & aug8        & 24.36 & 27.07 & 22.09 \\
          & combined256 & 21.65 & 24.32 & 23.31 \\
\bottomrule
\end{tabular}}
\end{table}

In the matched ablation, the single-pass model achieves slightly
lower greedy gaps than RELAGRAM at 10k and 20k updates (18.47/17.60\%
versus 18.89/17.86\%; Appendix~\ref{app:ablation}). Neither difference is
statistically significant: the comparison provides no evidence of an advantage
from recursion.

\paragraph{Recursion as refinement and search.}
RELAGRAM improves from its first stage to the trained fourth stage at every
size (Table~\ref{tab:depth}). Progress is non-monotonic: a later stage may
worsen before producing a better route, so the deepest output is not always
the strongest.
Applying the loop further yields gains that are negligible at
the trained size but grow with instance size: the best observed gap beyond depth four
improves on the depth-four gap by 0.05, 0.74, and 1.45 points on CVRP-100,
CVRP-500, and CVRP-1000 at depths six, eleven, and ten, respectively.

\begin{table}[t]
\centering\small
\caption{Greedy gap to PyVRP reference (\%; lower is better) by depth.
RELAGRAM trained at depth four, 160{,}000 optimizer updates.
$p$: two-sided paired $t$-test
against depth four ($\alpha=0.05$).}
\label{tab:depth}
{\fontsize{7.5}{8.5}\selectfont
\setlength{\tabcolsep}{4pt}
\begin{tabular}{lrrrrrrrcr}
\toprule
& \multicolumn{7}{c}{recursion depth} & & \\
\cmidrule(lr){2-8}
& 1 & 2 & 4 & 6 & 8 & 12 & 16 & best observed & $p$-value \\
\midrule
CVRP-100  & 15.29 & 13.69 & 13.40 & 13.35 & 13.75 & 14.69 & 15.49 & 13.35 @ 6  & 0.822 \\
CVRP-500  & 25.46 & 23.74 & 22.55 & 22.36 & 22.44 & 21.96 & 22.80 & 21.81 @ 11 & 0.015 \\
CVRP-1000 & 33.21 & 31.58 & 29.91 & 28.80 & 28.75 & 28.68 & 29.12 & 28.45 @ 10 & $<0.001$ \\
\bottomrule
\end{tabular}}
\end{table}

\section{Discussion and Conclusion}

RELAGRAM shows that a small latent-recursive model successfully learns
to route. It transfers without retraining to graphs up to ten times larger
and can improve beyond its trained depth, suggesting promising inference-time
depth scaling.
The model does not reach state-of-the-art results, and the matched greedy
comparison finds no significant advantage from recursion. We therefore
see RELAGRAM as a proof of concept for applying latent recursion
to graph combinatorial optimization.

We hypothesize that an important limitation is the sparse RL signal, which
needs further engineering for latent recursion. Importantly, $y$ and $z$
receive only indirect RL signal, making learning harder.

Lastly, RELAGRAM's recursive nature may be very useful for dynamic problems.
The loop already runs zero-shot past the conditions it was trained under; if it
handles an updated instance as well as extra depth, a single model with a decoder
could solve such problems end-to-end, by recurrently processing the last latent
state and the updated problem.

\paragraph{Acknowledgments.}
Thanks to the anonymous reviewers of the NeurIPS'26 MLxOR workshop.  This work was partially supported by the EFFILOAT project under the MOOI programme, funded by the Netherlands Enterprise Agency (RVO).

\bibliographystyle{unsrtnat}
\bibliography{reference}

\newpage

\appendix

\section{Further Model Evaluation}
\label{app:generality}
\label{app:ablation}

The main experiments compare RELAGRAM and DeepACO on solution quality. Here,
we look more closely at how the models work and provide a matched
comparison with a single-pass control.

Both models learn edge scores, but iteration happens in different places.
RELAGRAM repeatedly refines a latent state using the same network before decoding
the final heatmap. DeepACO instead uses an anisotropic GNN to produce a heuristic
for Ant Colony Optimization, where completed solutions update the pheromone state
and guide subsequent search. DeepACO performs better overall, although
RELAGRAM is competitive in two Combined256 settings: 17.11\% versus 17.38\% on
CVRP-500 and 24.32\% versus 23.31\% on CVRP-1000.

We compare RELAGRAM with a direct model that maps the
input straight to one heatmap. Both use the same 0.526M-parameter transformer
and are matched in optimizer updates, unique training instances, and sampled
routes. $p$-values are from two-sided paired $t$-tests over 256 shared
evaluation instances ($\alpha=0.05$).
The recursive/direct greedy gaps are 18.886/18.467\% at 10k updates
($p=0.226$) and 17.857/17.596\% at 20k ($p=0.406$). Neither difference is
statistically significant, so we find no evidence that RELAGRAM outperforms
this control under greedy decoding at these checkpoints.
RELAGRAM nevertheless provides an additional test-time scaling axis: greater depth can
improve solution quality without retraining, particularly after transfer to
larger instances.

\section{Inference-Time Scaling Details}
\label{app:inference}

We timed a single instance on the same 10\,GB A100 slice. One recursive step
takes 2.60, 36.32, and 131.40\,ms for $n=100,500,1000$; one greedy decode takes
36.39, 169.04, and 331.91\,ms. The recursive step is cheaper at every size,
although its cost rises on the larger transferred instances.

Aug8 is stronger than extra greedy depth: its gaps are 9.77\%, 17.61\%, and
24.36\%, versus 13.35\%, 21.81\%, and 28.45\% at the best tested depths. These
settings are not latency-matched: depth adds model passes but decodes one
route, while Aug8 decodes eight transformed instances. We therefore treat
depth as a new test-time computation axis.

\section{Self-Imitation Weight Sweep}
\label{app:si}

We also vary the self-imitation weight on CVRP-32, evaluating
$\lambda_{\mathrm{SI}}\in\{0,0.3,0.5,0.75,1\}$. Every run uses the same
1.707M-parameter model at 20k batches.

\begin{table}[h]
\centering\small
\caption{CVRP-32 gap to the PyVRP reference (\%; lower is
better).}
\label{tab:si-sweep}
\begin{tabular}{lrrrrr}
\toprule
$\lambda_{\mathrm{SI}}$ & 0 & 0.3 & 0.5 & 0.75 & 1 \\
\midrule
Greedy & 12.124 & 7.577 & 8.150 & 8.359 & 8.661 \\
Aug8   &  7.237 & 4.351 & 4.282 & 4.761 & 4.968 \\
\bottomrule
\end{tabular}
\end{table}

The model trains at both $\lambda_{\mathrm{SI}}=0$ and $1$, but performs best
with a mixed objective. $\lambda_{\mathrm{SI}}=0.3$ gives the strongest greedy
result, showing that self-imitation complements the route-level RL objective.

\end{document}